# Agentic Hazard Mapper: Emulating Expert Map-Making with AI Agents for Risk Communication

Sidi Wu
sidiwu@ethz.ch
ETH Zurich
Switzerland

Lifeng Lin
University of Tokyo
Japan

Yizi Chen
ETH Zurich
Switzerland

Anna Yokokubo
University of Tokyo
Japan

Lorenz Hurni
ETH Zurich
Switzerland

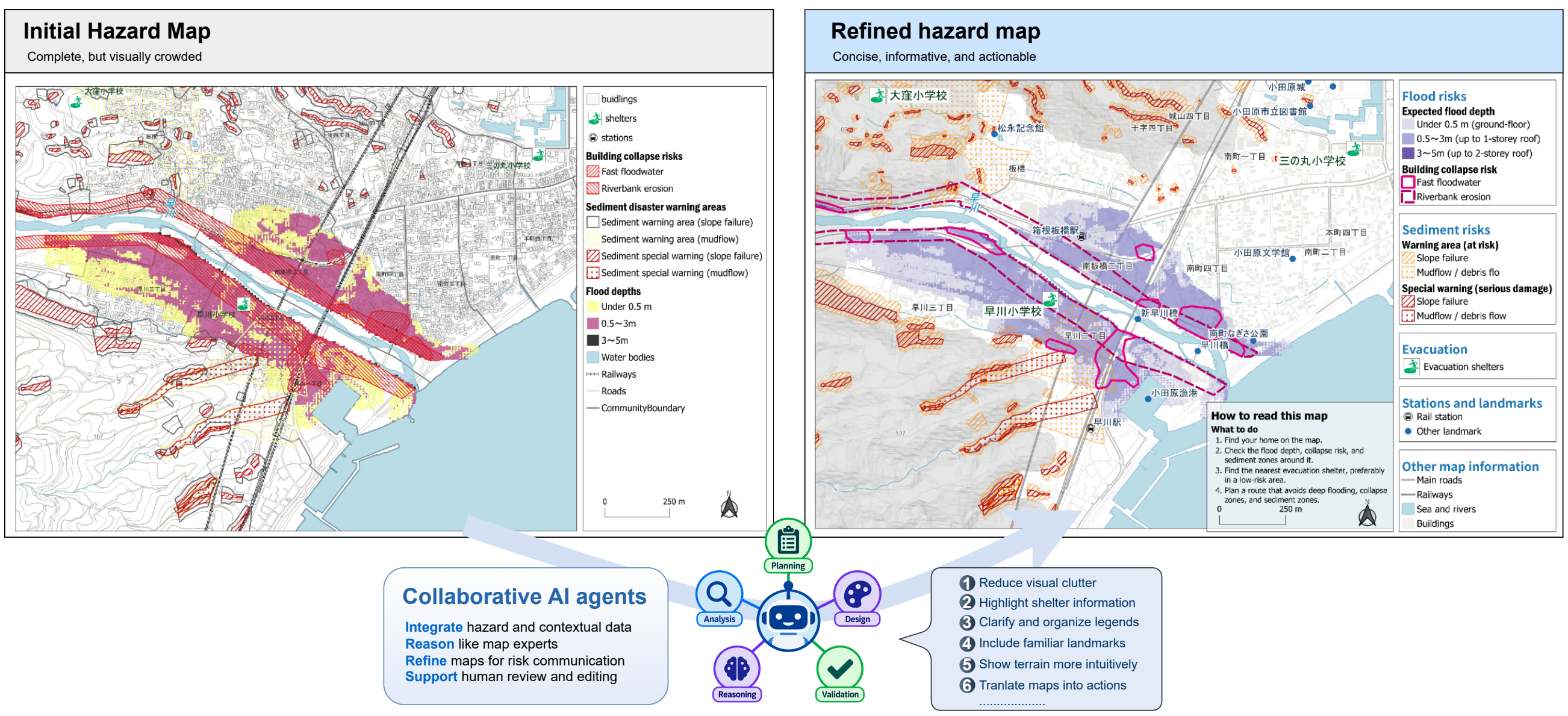

**Figure 1: We present *Agentic Hazard Mapper*, a multi-agent system that emulates cartographic expertise to transform complex hazard maps into clearer, more actionable representations for risk communication. By coordinating specialized AI agents across stages of the professional cartographic workflow, the system emulates the iterative reasoning and design trade-offs of expert map-making, particularly in contexts where cartographic expertise is limited.**

## Abstract

Hazard maps are essential tools for public risk communication, preparedness, and evacuation planning, yet effective cartographic design is not always ensured due to limited access to cartographic expertise across local governments and communities. We present *Agentic Hazard Mapper*, a multi-agent system that automates hazard-map production by emulating professional cartographic workflows and iterative design trade-offs while adapting design decisions to different data, locations, and hazard contexts. We conducted a formative study with map experts to inform the system design, followed by a user study and interviews with map experts and hazard practitioners. The generated maps improved participants' spatial understanding and evacuation-planning performance, reduced perceived workload and task completion time, and were generally preferred over corresponding municipal maps. We further discuss human oversight, adaptation across users and communication contexts, and opportunities to extend agentic support across broader stages of hazard-map production and risk communication.

## 1 Introduction

Natural hazards can cause substantial loss of life, damage to infrastructure, and disruption to communities. Japan is particularly susceptible to a wide range of natural hazards, including earthquakes, tsunamis, floods, and landslides, which occur across the country on a recurring basis [4]. In recent years, large-scale disasters have occurred, such as the 2016 Kumamoto Earthquake, Typhoon No. 14 in 2022, and heavy rainfalls caused by the Baiu Rainy Season

Front in 2023. To mitigate the damage caused by natural hazards, effective risk communication is necessary for the general public to understand and prepare for the hazards. Hazard maps offer an important instrument for risk communication, as they present spatial information and intensity of potential hazards, which help people quickly gain knowledge about hazard extent, evacuation routes, and local terrain, compared to text-only messages [23, 25]. As such, hazard maps must ensure that the general public can understand and correctly interpret the presented information [25].

However, hazard-map production in Japan is decentralized across more than 1,700 municipalities, where the access to professional cartographic expertise can be limited or uneven, leaving many design decisions to practitioners whose primary expertise lies in hazard analysis or disaster management rather than cartography and visualization. This results in considerable variation in map design and generally limited attention to cartographic representation [26]. Such variation can substantially affect how people interpret hazard information and, in some cases, increase the likelihood of misinterpretation [27]. This gap is also reflected in residents' experiences with hazard maps. According to a nationwide survey conducted in 2021, only 23.2% of respondents who had viewed their municipal hazard map considered it *very useful* for evacuation planning, while 56.4% considered it somewhat useful [30]. Respondents who found the maps less useful reported difficulties in map reading such as identifying the type and severity of hazards affecting their homes, deciding on the evacuation destination, and determining which evacuation routes to take. These responses highlight persistent challenges in translating technical hazard information into visual representations that support effective risk communication.

Addressing these challenges calls for an automated system that can incorporate cartographic expertise in contexts where such expertise is limited. At the same time, rather than applying a single uniform design standard, such a system should remain flexible enough to make context-dependent adaptations, as data availability, hazard characteristics, geographic conditions, and local communication needs vary considerably across municipalities. These requirements are difficult to capture in a static workflow, where design decisions are predefined. Recent advances in AI provide new opportunities for supporting this type of adaptive reasoning. Agentic AI is particularly promising because it can address more open-ended and context-sensitive problems beyond narrow, well-defined tasks. Such systems coordinate specialized large language model (LLM) agents, decompose complex tasks into multiple stages, dynamically adapt strategies to changing conditions, and iteratively refine their decisions based on intermediate results and feedback [6, 10, 40]. These capabilities make agentic systems well suited to hazard map-making.

In this paper, we propose an agentic system for automatically producing hazard maps for effective risk communication. We focus on printed maps, as they remain one of the most widely used media for hazard communication in Japan. The system is designed to iteratively refine map quality, adapt to different data, locations, and hazard characteristics, and remain grounded in established cartographic practices, rules, and preferences. We use insights from a formative study with map experts to inform the system design, and then conduct a user study as well as expert interviews to investigate how the cartographic expertise is operationized through the system, how the resulting maps are perceived, and how such a system could be used in actual hazard-preparedness contexts.

To summarize, our contributions are:

- We conducted a formative study with map experts to characterize expert map-making practices, including explicit cartographic guidelines and tacit design preferences for effective visual communication, and used these insights to inform the design of our agentic system.
- We designed and implemented an agentic map-making system that incorporates expert cartographic knowledge and practices to automatically generate hazard maps, with the goal of supporting map production in contexts where cartographic expertise is limited or unavailable.
- We conducted a user study to evaluate how effectively the generated maps support spatial understanding, risk comprehension, and evacuation planning.
- We conducted interviews with map experts and hazard practitioners to assess the cartographic quality and practical utility of the generated maps, and to identify opportunities and challenges for integrating agentic map-making systems into real-world hazard-preparedness workflows.

## 2 Related work

### 2.1 Risk communication and hazard maps

Broadly speaking, there are two types of hazard risk communication: warnings of incoming threats and communicating long-term risks for early preparedness and mitigation [39]. For the latter, maps have long been recognized for pre-disaster preparation and evacuation route planning [23]. However, the visual communication of maps is not always effective. For example, for the case of hurricane forecasting maps, the general public can sometimes struggle with interpreting geographic risk areas or uncertainty levels [3]. More generally, substantial variation exists in how people comprehend hazard maps and draw appropriate conclusions from the spatial information presented [22]. Recent reviews [48] continued to highlight challenges related to readability, visual complexity, color conventions, and risk perception.

The effective visual design of hazard maps has long been studied in cartography and risk communication. Prior work [28] has examined how flood-risk communication through maps can increase awareness and support preparedness. Other works have investigated how hazard information can be represented with an appropriate balance between simplicity and complexity [13], how cartographic variables such as color, symbols, size, typography, and basemap styles influence the readability and interpretation of flood and evacuation maps [12, 21, 27, 41]. Works like [21, 34] have investigated cartographic recommendations and explicit design criteria for effective risk communication for flood and seismic hazards. Rather than a uniform design scheme, Meyer et al. [29] proposed user-specific adaptations in map content and visualization. Beyond visual design, Koesten et al. [20] showed that crisis-map interpretation can also be hindered by insufficient explanatory information, limited personal relevance, and distrust in the visualization. Together, these studies provide an established body of cartographic knowledge for hazard maps and inspire the design of our system.

Communication of imminent hazards, in contrast, often extends beyond maps alone and combines textual warnings, alerts, visual information, and other modalities to support timely protective action [23]. Prior work has examined how emergency organizations communicate with the public through online and social-media channels during emerging disasters [17], as well as how technologies can support professionals in authoring effective severe-weather risk messages [18]. Such warning scenarios involve rapidly changing conditions, greater temporal pressure, and immediate protective actions. In this work, we focus primarily on map-based communication for longer-term hazard preparedness and evacuation planning, where cartographic design can support spatial understanding and planning before an emergency unfolds.

## 2.2 LLMs and agentic systems for automated visualization

Automating visualization design has long been a goal of the visualization community. Traditional approaches typically formalize visualization knowledge as explicit rules, design guidelines, or optimization constraints, such as Voyager [45], Draco [31], and VizML [16]. While these approaches can produce effective visualizations, they often rely on predefined design spaces and are less flexible in handling complex natural-language requests, ambiguous user intentions, and diverse data characteristics.

Recent advances in generative AI have enabled a new paradigm for automated visualization. Early approaches such as Data2Vis [9] learned visualization specifications from a given dataset, which are later transformed to visualization plots. More recently, LLMs have extended this paradigm by interpreting natural-language requests, reasoning about visualization choices, and generating executable visualization code. They been applied to various visualization tasks, including visualization recommendation and design assistance [36, 44], visualization code generation [5, 50], and visualization evaluation and critique [37]. Compared with traditional rule-based systems, LLMs provide greater flexibility in interpreting user requests and synthesizing visualization designs from heterogeneous sources of knowledge. Beyond visualization, LLM-based agents have demonstrated the ability to perform complex planning and decision-making tasks in open-ended environments. For example, [43] introduced an embodied LLM agent for Minecraft that continuously acquires skills, interacts with the environment, and accumulates knowledge through iterative exploration.

More recently, researchers have explored multi-agent systems in which multiple specialized LLM agents collaborate to solve complex tasks. Such systems have been applied to code generation [15], software engineering [32], and general-purpose agent frameworks [46]. Rather than relying on a single model, multi-agent systems decompose problems into specialized subtasks and coordinate the reasoning processes of multiple agents, often leading to improved robustness, planning, and error recovery. Within visualization research, several agentic systems have recently emerged such as MatPlotAgent [47], CoDA [6], and VisPath [35]. Given user requirements in natural language and input datasets, these systems can analyze tasks, select visualization strategies, generate executable code, verify outputs, and iteratively refine results. For example, CoDA [6] adopts a collaborative multi-agent architecture consisting of specialized agents for metadata analysis, task planning, code generation, debugging, and self-reflection. Our agentic map-generation system can be situated within the broader landscape of LLM-powered visualization and multi-agent systems.

## 2.3 AI-assisted map-making

Early AI-assisted map-making primarily explored map generation and style transfer through image synthesis. These approaches include GAN-based methods [7, 19], diffusion-based generation [1], and direct use of off-the-shelf models such as DALL·E [8, 11]. Although these methods can produce visually compelling results, pixel-based generation may introduce hallucinated or geographically inaccurate content and offers limited flexibility and fine-grained control over cartographic elements during map generation.

More recently, LLMs have been integrated with GIS and cartographic tools to support more structured map-making workflows. MapGPT [49] uses an LLM to interpret natural-language mapping requirements, decompose user requests into subtasks, and invoke specialized cartographic tools to manipulate individual map elements. By grounding the generation process in structured geographic data and tool-based operations, MapGPT provides greater control over map content and reduces the risk of geographic hallucinations. While MapGPT uses a single, centralized agent for map generation, CartoAgent [42] employs a multi-agent framework for map styling and restyling, consisting of preparation, design, and evaluation agents. The preparation agent extracts style information, such as color palettes, from user-provided reference maps; the design agent proposes color schemes and map symbols; and the evaluation agent assesses whether the generated map is visually effective and informative. The design is iteratively revised until it satisfies the evaluation criteria. Its workflow is mainly structured as a staged pipeline for map styling and evaluation, with limited emphasis on persistent memory, targeted reflection, or richer inter-agent collaboration across the broader cartographic workflow. Building on this trajectory, our system models a broader cartographic cycle spanning data processing, design and layout, evaluation, and revision, with specialized agents grounded in cartographic knowledge, practices, and expert preferences.

# 3 Formative Study

To understand professional map-making practices and the challenges surrounding current hazard-map visualization in Japan, we conducted a formative study with map experts. We focused on how experts approach the map-making process, the cartographic principles and constraints that guide their decisions, the challenges they perceive in existing hazard maps, and the evaluation criteria to assess the map quality, in the context of hazard mapping and risk communication. The goal of the study was to elicit both explicit design knowledge and tacit professional judgments that could inform the design of our agentic hazard-mapping system.

## 3.1 Method

We first reviewed literature on cartographic design, with particular attention to hazard mapping and risk communication. From the literature [12, 13, 21, 24, 27, 33, 34, 38, 48], we extracted a set of fundamental cartographic principles and identified 16 criteria describing what an effective hazard map should achieve and how they could be operationalized through cartographic variables such as hue, lightness, symbol size, line style, texture, and layer order. Requirements prescribed by official hazard-mapping guidelines were treated separately as fixed constraints.

We recruited five experts (E1–E5) with between 3 and 40 years of professional experience in cartography, including two with particular expertise in hazard visualization. During structured, on-site interviews, experts first described their typical cartographic workflows. They then assessed three existing Japanese hazard maps with respect to cartographic principles, risk communication, and opportunities for improvement. Finally, experts evaluated the 16 literature-derived criteria, ranked their importance, selected the eight they considered most critical, and explained their choices. Afterward, they discussed how the design variables of these criteria should be adapted to different hazard types, data complexity, geographic characteristics, and communication needs.

## 3.2 Insights and findings

**Cartographic Workflow** Experts described map-making as an iterative rather than linear process. A typical cartographic workflow starts by deciding determining the map's purposes and intended audience, followed by collecting and checking data, selecting relevant information, and establishing a visual hierarchy to decide what important features stand out from the background. Cartographers then design colors, symbols, labels, and layout, scrutinize the result for mistakes or confusing details, and revise their earlier decisions as needed before publication.

**Problems in Current Hazard Maps** The experts identified five main issues in current hazard maps: ***1) Information overload.*** All experts noted that hazard maps often retain excessive geographic detail, including dense roads, buildings, labels, and contextual features that are not always relevant to the communicated hazard. Such information can compete with critical risk information and reduce readability. Experts also noted overly detailed classifications (E1, E2, E4) or legends containing unnecessary or unused categories (E3), which can increase users' interpretive burden: ***2) Weak visual hierarchy and confusing visual encoding.*** E3 and E5 observed that hazard-related information is not always visually distinguished from basemap information used primarily for orientation and context. They also noted confusing visual encoding: multiple hues within the same hazard type can make it difficult to perceive the ordering of severity levels, while similar colors or fill patterns across hazard types can make different hazards difficult to distinguish. ***3) Limited interpretability of legends and symbols.*** E1, E2, and E3 noted that legends often lack clear structure and provide limited explanation. Rather than relying heavily on extensive legends, E3, E4, and E5 suggested using more intuitive color schemes and self-explanatory symbols can provoke users' existing visual associations and reduce the need for repeated legend consultation. ***4) Insufficient emphasis on meaningful labels.*** E4 and E5 emphasized the importance of labels for shelters, communities, landmarks, and other recognizable locations that help residents orient themselves and plan evacuation routes. Such information should be selectively emphasized rather than treated as equivalent to less relevant geographic annotations. ***5) Inadequate representation of topography and elevation.*** Most experts (E1, E3, E4, and E5) emphasized the importance of terrain and elevation particularly for floods and landslides. However, this information is either absent or difficult to interpret in existing maps. For example, conventional contour lines can convey precise elevation information but may be less intuitive for the general public. Experts also identified useful existing practices. E1, E2, and E3 valued illustrations linking flood-inundation depths to affected building storeys, as they make abstract depth values easier to interpret. E2 and E4 also noted that historical hazard information can strengthen risk awareness by connecting risks to past experience, although it is less useful for assessing current conditions.

**Aspects for Effective Risk Communication** Across the interviews, four broader communication goals emerged. First, hazard maps should support *risk awareness* by making affected areas and relative severity readily perceivable (E1–E5). Second, they should support *spatial awareness* by helping users identify shelters, neighborhoods, roads, terrain, and potential evacuation routes (E1–E5). Third, maps should establish *personal relevance*, allowing users to understand how hazards may affect their own circumstances and possible actions (E1–E5). Finally, E3 and E5 emphasized *multi-hazard awareness*, to distinguish and associate hazards that overlap spatially or interact in ways relevant to evacuation decisions.

**Quality criteria for hazard map design** Experts prioritized eight criteria for guiding design and evaluation: *information density*, *hazard differentiation*, *visual hierarchy*, *symbol interpretability*, *basemap clarity*, *orientation and localizability*, *label readability*, and *legend clarity*. Information density received particularly strong emphasis, with all experts advocating the simplification or removal of less relevant information when it competes with hazard-related content. Experts also emphasized clear differentiation between hazard types and severity levels through appropriate color encoding and intuitive symbols. They also emphasized a strong visual hierarchy that can be established through variables such as size, font, color, and transparency. At the same time, basemap information should remain sufficiently legible to support orientation, particularly major roads and topographic features. Labels important for localization and evacuation, such as shelter names, community names, landmarks, and facility names, should remain identifiable and unobscured by competing visual elements. Finally, legends should be concise, clearly organized, and use plain-language descriptions where possible. By contrast, criteria related primarily to aesthetics and color harmony received lower priority. We derived weights for each criterion based on how many experts have voted for this. Importantly, experts stressed that these criteria cannot be applied through a single fixed design scheme. Their implementation depends on factors such as map scale, spatial extent, available data, terrain, and hazard characteristics. Design choices can also conflict with one another, requiring trade-offs and iterative adjustment. This makes flexible professional judgment central to effective hazard-map design.

## 3.3 Design implications

Based on these findings, we made the following design decisions for our system:

**Risk-centered information organization and visual hierarchy.** The system should organize and simplify information according to the communication goal, prioritizing content that directly supports risk understanding and evacuation planning while secondary information can be reduced or omitted when appropriate. This priority should also be reflected visually, with the dominant hazard visualization and still sufficiently legible contextual information to support orientation and localization.

**Adaptive and flexible approach.** Rather than applying a uniform visual template, the system should adapt to hazard type and severity, available data, terrain, urban structure, map scale, and locally relevant landmarks or evacuation resources. Established criteria and expert heuristics should guide decisions without prescribing a single fixed solution, thereby preserving flexibility in selecting appropriate visual variables, parameters, and representations.

**Semantically and spatially grounded representation.** Colors, symbols, labels, and other visual variables should preserve clear semantic distinctions between hazard types, severity levels, and contextual information. At the same time, maps should preserve meaningful spatial anchors for orientation and localization, and provide interpretable terrain information when topography is relevant to risk understanding.

**Iterative and multi-objective map-making.** Hazard-map design should be treated as an iterative, multi-stage process in which decisions made at one stage can affect later stages and improvements in one quality dimension may conflict with another. The system should therefore evaluate intermediate results, coordinate decisions across stages, and iteratively revise the resulting map to balance competing cartographic objectives.

# 4 The Agentic Hazard Mapper

## 4.1 System Overview

The overview of our proposed agentic system is shown in Figure 2. Mimicking real-world cartographic practice, we design multiple specialized agents, each focusing on a well-defined area of expertise. Given the input data from various sources, the *Data agent* summarizes, selects, processes and prioritizes the data based on their relevance to the target hazard. The *Design agent* determines visual encoding for each data layer, while the *Layout agent* organizes the map layout, legends, and supplementary explanations. Each agent produces a lightweight configuration file that can be easily passed to other agents and to the renderer to generate maps. The rendered map is then accessed by the *Evaluation agent* across multiple quality dimensions according to expert-derived cartographic criteria and provides feedback for refinement. The *Revision agent* interprets the evaluation feedback and determines the appropriate improvements for the corresponding upstream agents. The system iteratively regenerates and evaluates the map until the predefined acceptance criteria are satisfied or the maximum number of iterations is reached.

## 4.2 Knowledge and Constraint Grounding

Based on the formative study and collected cartographic and hazard-management documents, we derive a set of hard and soft constraints to guide map generation. *Hard constraints* capture requirements that must always be satisfied. These include fundamental cartographic rules, such as appropriate visual encoding for different geometry types and the inclusion of essential map elements, as well as official hazard-mapping requirements. For example, the geographical extent, hazard zones, hazard categories, and their meanings must remain unchanged throughout the generation process. *Soft constraints* capture cartographic recommendations, design heuristics, and common mapping practices that allow for contextual interpretation and flexibility. They include considerations such as adapting color schemes to local characteristics, determining the appropriate level of background detail when hazards overlap with other features, and incorporating terrain information when it is relevant for risk communication. Rather than prescribing a single design solution, these constraints provide contextual knowledge and recommendations that guide design decisions. Within our multi-agent framework, hard constraints are enforced through deterministic operations and validation checks that prevent violations, whereas soft constraints are encoded as system instructions and recommendations that inform the reasoning and decision-making of individual LLM agents.

## 4.3 Data Agent

Given geospatial data from heterogeneous sources, such as hazard information, building footprints, street networks, rivers, shelter locations, points of interest (PoIs), community names, and terrain characteristics, the *Data agent* first extracts lightweight metadata summaries from the input files using Python-based geospatial tools. These summaries describe properties such as geometry types, statistics, spatial distributions, and relevant attributes. The agent then assesses the relevance and importance of each data layer with respect to the target hazard and recommends appropriate data-processing operations based on the amount and characteristics of the available information. The operations include:

- **Layer selection:** remove irrelevant or redundant layers.
- **Feature filtering:** retain only salient or representative features, such as major roads, large buildings, or well-known landmarks.
- **Prioritization:** estimate the relative importance of layers and features for risk communication, such as prioritizing shelters for flood preparedness.
- **Ordering:** determine the rendering order across layers and among features within a layer based on geometry type, semantic importance, and relevant attributes. For example, polygon layers are typically rendered beneath line and point layers, while features within a layer may be ordered by size, class, or importance to reduce occlusion and maintain visual hierarchy.

The resulting data summaries and processing decisions are stored explicitly in a structured JSON configuration, allowing downstream agents to reason over both the selected information and the transformations applied to it.

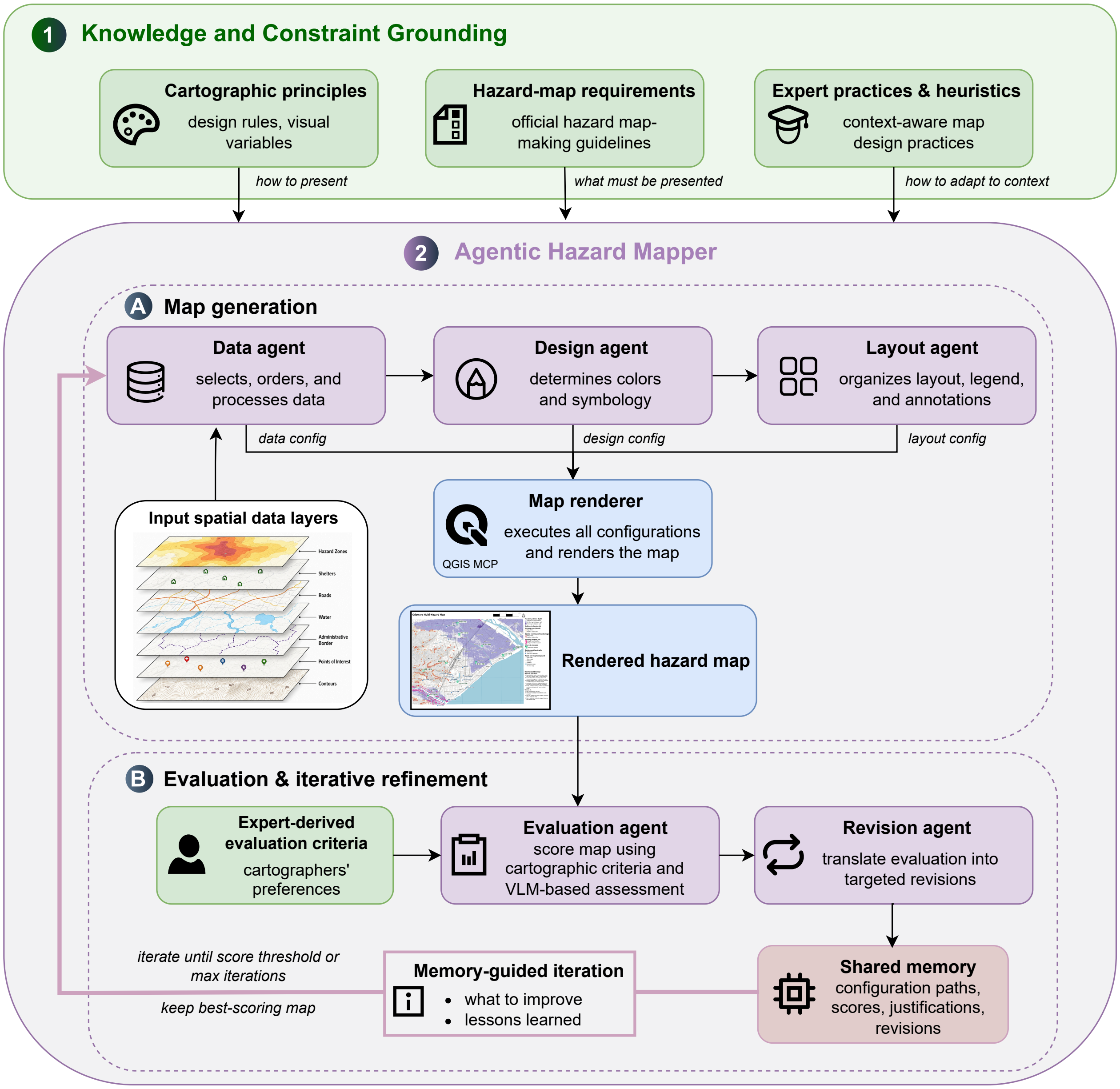

**Figure 2: The overview of our proposed Agentic Hazard Mapper. Our system contains multiple specialized agents for data processing, design, layout, evaluation, and revision, grounded in cartographic knowledge and hazard mapping constraints. Our system follows an iterative process until it reaches the score threshold or maximum iterations.**

## 4.4 Design Agent

The *Design agent* receives the configuration produced by the *Data agent* and determines the design operations, which include:

- **Visual encoding:** assign visual properties such as color, symbology, size, line width, typography, and transparency to each data layer and their subclasses according to geometry type, semantic meaning, and category type such as nominal, ordinal or quantitative.

- **Compound encoding:** combine multiple visual channels to represent different attributes simultaneously. For example, color intensity may encode hazard severity or importance, while symbol shape distinguishes semantic categories.
- **Cross-layer coordination:** coordinate visual styles across layers and subclasses to maintain semantic consistency while preserving perceptual distinction. For example, features belonging to the same semantic family may share a common base color, while subcategories are differentiated through variations in lightness, saturation, shape, or size.

The resulting visual-design decisions for each data layer are stored in the shared JSON configuration.

## 4.5 Layout Agent

The *Layout agent* receives the visual design decisions from the *Design agent*, together with the original data characteristics and priorities identified by the *Data agent*, and determines the organization of the map. Its decisions include:

- **Legend organization:** group and arrange legend items according to their semantic relationships, topic, and importance.
- **Legend refinement:** improve the clarity and readability of legend labels and descriptions using concise, plain language.
- **Explanatory content:** add and arrange supplementary panels with annotations that explain key map elements, hazard information, or recommended interpretations.
- **Layout adjustment:** adjust spacing, text wrapping, and the placement of map components to reduce unused space and avoid crowding, particularly when adapting the layouts to languages with different text lengths and writing characteristics, such as English and Japanese.

The layout decisions, together with the refined legend labels and explanatory text, are stored in a shared JSON configuration file.

## 4.6 Renderer

The *Renderer* takes the configurations produced by the *Data agent*, *Design agent*, and *Layout agent* and executes the specified data-processing, visual-design, and layout operations in QGIS[1], a widely used open-source platform for processing, analyzing, and visualizing geospatial data and producing maps. The renderer interacts with QGIS through the Model Context Protocol (MCP), which exposes QGIS functionalities as callable tools for programmatically manipulating geospatial data, applying cartographic styles, arranging map elements and layouts, and exporting the final map image. Unlike the LLM-based agents, the renderer does not make independent cartographic decisions. Instead, it deterministically executes the operations specified in the shared configurations, providing a clear separation between agent reasoning and map rendering.

[1]https://qgis.org/

## 4.7 Evaluation Agent

Given a rendered map image, the *Evaluation agent* uses a vision-language model (VLM) to assess map quality according to evaluation criteria derived from expert input. To reduce the ambiguity from broad description and improve the consistency of VLM-based evaluation, each criterion is decomposed into a set of explicit sub-conditions. The agent evaluates whether each sub-condition is satisfied and aggregates these judgments into a criterion-level score. For example, *basemap clarity* considers whether contextual layers such as buildings and roads remain visually subordinate to hazard information, whether they can be distinguished from hazard layers, and whether important structures remain legible beneath hazard overlays. The complete evaluation rubric is provided in Appendix **??**.

The overall quality score is computed as a weighted combination of the individual criteria, with weights derived from expert input collected during the formative study. In the first iteration, the system evaluates the rendered map independently and produces an absolute score for each criterion together with a justification. In subsequent iterations, the *Evaluation agent* compares the current map with the previous generation and estimates the change ($\Delta$) for each criterion. This pairwise comparison helps mitigate variability in VLM judgments across iterations by focusing on relative improvements or degradations between successive generations, rather than relying solely on independently calibrated absolute scores. For each criterion, the score at iteration $t$ is updated as $S_t = S_{t-1} + \Delta_t$, where $\Delta_t$ denotes the VLM-estimated change relative to the previous generation. The criterion-level scores and their corresponding justifications are stored in a single configuration file for each iteration.

## 4.8 Revision Agent

Given the feedback from the *Evaluation agent*, the *Revision agent* decomposes identified issues and routes them to the corresponding upstream agents for refinement. For example, if different hazard classes are not sufficiently distinguishable, the *Revision agent* directs the issue to the *Design agent*; if the legend or explanatory panels are overly text-heavy, the feedback is instead routed to the *Layout agent*. Similarly, issues related to excessive map content are assigned to the *Data agent*. To prevent the *Revision agent* from proposing changes that cannot be executed by the system, the supported capabilities of each upstream agent are explicitly provided as part of its context. This ensures that revision suggestions remain actionable within the capabilities of the current system.

## 4.9 Multi-Agent Coordination and Iterative Refinement

Beyond the responsibilities of individual agents, the system maintains a shared state that enables agents to build on earlier decisions and coordinate across stages. Within each generation cycle, agents are executed in a sequence: *Data Agent* → *Design Agent* → *Layout Agent* → *Evaluation Agent* → *Revision Agent*. The outputs of upstream agents are written to the shared state that stores their configurations as structured JSON files and become part of the context available to downstream agents, allowing later decisions to remain consistent with earlier ones.

At the end of each iteration, the system records criterion-level evaluation scores, corresponding justifications, and revision suggestions targeted at specific agents. These records are accumulated in shared memory and summarized before the next iteration into two forms of guidance:

- **What to improve.** The system identifies the iteration with the highest overall evaluation score and summarizes its four highest-impact weaknesses together with the corresponding revision suggestions and responsible agents. This allows subsequent iterations to build on the strongest previous result rather than restarting from the most recent output.
- **Lessons from prior attempts.** The system summarizes changes across previous iterations and identifies revisions associated with increases or decreases in evaluation scores. This provides a history of prior attempts and helps reduce oscillation, such as repeatedly introducing and reverting the same design changes.

These summaries are passed to the relevant upstream agents as contextual guidance for the next generation cycle. Distilling the interaction history in this manner reduces redundant contextual information and token consumption while retaining the most actionable information from previous attempts. The system continues the generation–evaluation–revision cycle until either the predefined acceptance criteria are satisfied or the maximum number of iterations is reached. If the stopping threshold is not reached, the highest-scoring map generated across all iterations is retained as the final output. For the individual agents, we use Claude Sonnet 4.6 for the Data Agent, Claude Opus 4.8 for the Design Agent, and GPT-5.5 for the Layout, Evaluation, and Revision Agents.

## 5 User Study

To evaluate the effectiveness of risk communication in the maps generated by our system, we conducted a user study with 24 participants. We examined whether the generated maps support spatial and hazard understanding, facilitate evacuation planning, reduce perceived workload and task-completion time, and are preferred over the corresponding municipality-produced hazard maps.

### 5.1 Method

**Participants.** We recruited 24 participants (15 male, 9 female; ages 21-31, mean=25) who has been living in Japan for more than 1 year. All participants have minimal or no prior experience with hazard mapping and risk communication.

**Study design.** We employed a within-subjects experimental design to compare two conditions: an original municipality map and a map generated by our AI system. Each participant experienced both conditions using maps from different study areas within the same municipality to reduce memorization from repeated exposure to the same location. Condition order was counterbalanced across participants.

**Task and materials.** We selected hazard maps from two Japanese municipalities representing different geographic and hazard contexts: Odawara, characterized by mountainous terrain and compound exposure to river flooding and sediment-related hazards, and Ube, a more densely populated coastal city with prominent storm-surge risk. The two municipalities have different flood mechanisms and types. Twelve participants were assigned to each municipality, and two administrative divisions were selected within each municipality for the two experimental conditions. For both conditions, we prepared A1-sized PDF maps displayed on a 14-inch MacBook and provided an A4 printout for drawing tasks. Participants could freely zoom and pan the digital maps up to the full-resolution A1 output, ensuring comparable access to map detail across conditions. Participants completed a set of map-reading tasks designed to assess four dimensions identified in our formative study:

- *Spatial awareness:* locating communities, hazard zones, and shelters; comparing elevation; and identifying routes between specified locations.
- *Risk awareness:* identifying hazard types and affected areas and comparing risk severity across locations.
- *Personal relevance:* determining risks affecting a hypothetical home location, identifying an appropriate shelter, and planning an evacuation route.
- *Multi-hazard awareness.* identifying multiple hazards affecting the same area and reasoning about their implications for evacuation.

**Procedure.** Each study session lasted approximately one hour and was conducted in person. Participants received a small gift as compensation. At the beginning, participants were given a brief explanation of the study purpose before they filled out the informed consent form. Participants then completed a pre-study questionnaire on demographics, prior familiarity with hazard maps, and perceptions of existing hazard maps Next, they completed the map-reading tasks under two map conditions. We recorded completion time for each condition, and participants completed the NASA-TLX questionnaire after each condition. Finally, in the post-study questionnaire participants rated both map conditions on 7-point Likert scales, indicated their overall preference, and provided open-ended feedback on perceived strengths, limitations, and opportunities for improvement.

### 5.2 Results

**Hazard familiarity and perception.** According to the pre-study questionnaire, most participants had some prior familiarity with hazard maps: 10 reported being *moderately familiar*, 10 *slightly familiar*, and four *not at all familiar*. Among the 20 participants with some familiarity, most had encountered flood hazard maps, while approximately half were also familiar with tsunami or earthquake hazard maps. Despite this prior exposure, 21 of the 24 participants reported that they rarely or never used hazard maps in practice. Participants also described several challenges with existing hazard maps. These included difficulty understanding the maps (3 participants), limited prior instruction in school on how to use them (1), difficulty locating useful information because of excessive detail (3), and uncertainty about how to translate mapped information into concrete actions or decisions (4). When asked what makes a hazard map effective for risk communication, *ease of understanding* was mentioned most frequently (10 participants). Related responses emphasized simple design, clear legends, and underscored risk representation (3), while others highlighted personalization to individual needs (2) and online or real-time visualization (3).

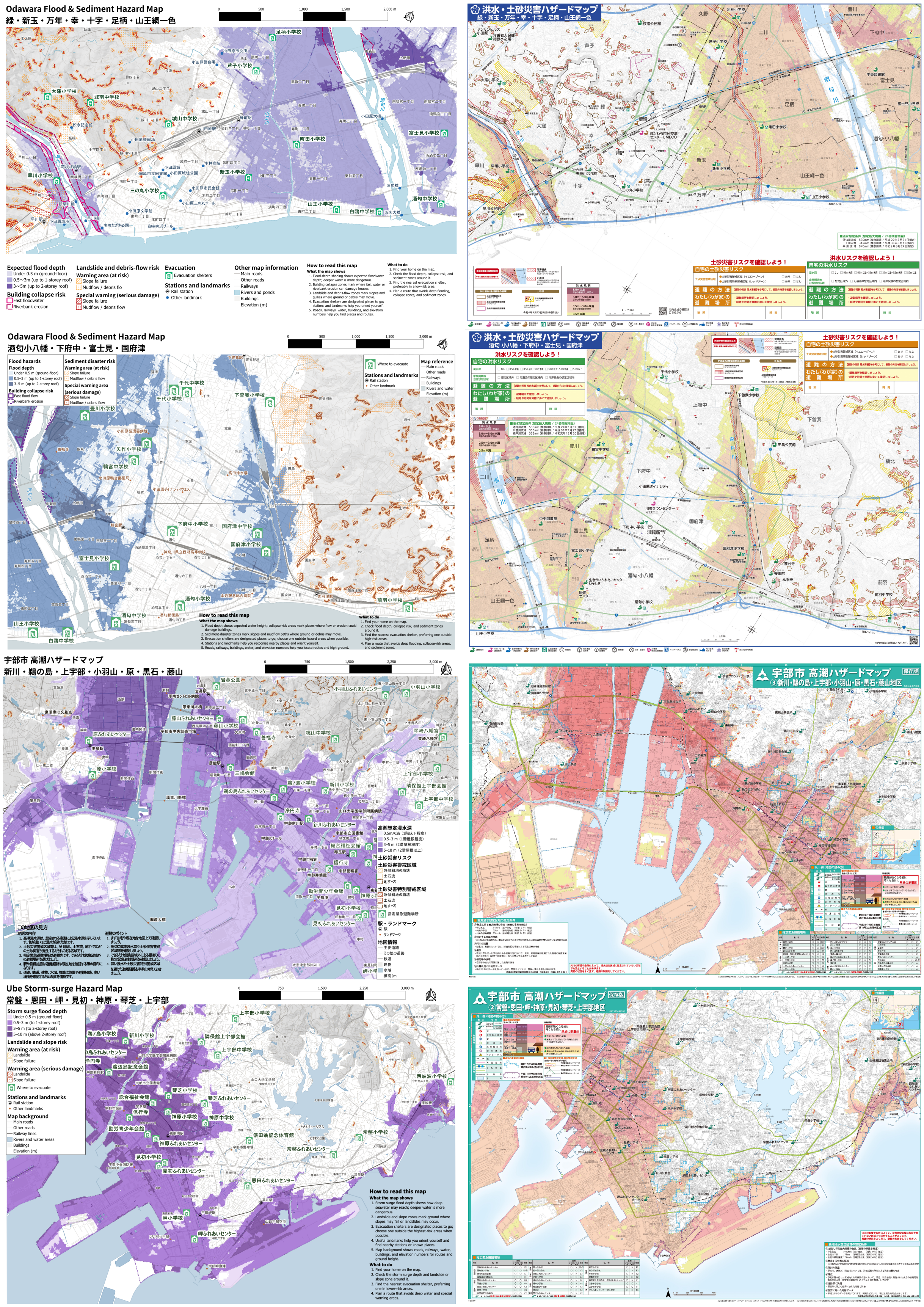

**Figure 3: Maps used in the user study. Left: system-generated map, with the layout post-adjusted to match the official map; right: corresponding official municipality map.**

**Accuracy and completion time.** We computed an overall accuracy score by weighting the four evaluation dimensions equally. Within each dimension, scores for individual questions were normalized to the range $[0, 1]$ and then averaged. Single-choice questions were scored using a binary scheme. For route-planning tasks, we evaluated whether the general route was correct, whether it followed the road network, and whether the required turns and intersections were correctly identified. For open-ended reasoning questions, responses were scored according to whether they demonstrated sufficient awareness of the relevant hazards and risks. The four dimension-level scores were then averaged to obtain an overall accuracy score for each participant and condition. We analyzed accuracy and completion time using linear mixed-effects models, with map condition, study area, flood type, condition order, and task order as fixed effects and participant as a random effect to account for repeated measures. As shown in Figure 4, participants achieved significantly higher overall accuracy when using maps generated by our system ($M = 0.75$, $SD = 0.10$) than when using the official municipality maps ($M = 0.66$, $SD = 0.09$). The mixed-effects model showed a significant main effect of map condition ($\beta = 0.085$, $p < .001$), indicating an approximately 8.5-percentage-point improvement in accuracy with our maps. No significant main effects of other covariates like study area, condition order, task order, and flood type were observed ($p > .05$). Participants also completed the tasks significantly faster with our maps ($M = 19.89$ min, $SD = 4.85$) than with the official maps ($M = 26.51$ min, $SD = 5.74$). The mixed-effects model showed a significant effect of map condition on completion time ($\beta = -6.61$ min, $p < .001$). We additionally observed a significant effect of task order ($\beta = -5.67$ min, $p < .001$), suggesting a practice effect whereby participants completed their second task faster regardless of map condition. Importantly, the effect of map condition remained significant after accounting for this order effect.

**User Experience and Cognitive Load.** We normalized the NASA-TLX ratings to the range $[-5, 5]$, with lower values indicating lower perceived workload. As shown in Figure 5, participants reported significantly lower workload when using the maps generated by our system than when using the official municipality maps. Significant reductions were observed for *Mental Demand* ($\beta = -1.46$, $p < .001$), *Physical Demand* ($\beta = -2.65$, $p < .001$), *Effort* ($\beta = -1.83$, $p < .001$), *Frustration* ($\beta = -1.96$, $p < .001$), and *Temporal Demand* ($\beta = -0.90$, $p = .015$). These results suggest that the system-generated maps reduced the effort required to locate and interpret relevant information, supported more efficient map reading, and resulted in a less frustrating user experience. In contrast, we found no significant difference in participants' perceived performance between the two conditions ($\beta = -0.09$, $p = .903$). This is consistent with the absence of a significant difference in participants' reported confidence between the two map conditions. We also found no significant main effects of flood type, study area, presentation order, or task order across the NASA-TLX subscales ($p > .05$).

**User preference.** In the post-study questionnaire, participants evaluated the two map conditions across seven aspects: Q1, *map clarity and readability*; Q2, *appropriateness of information amount*; Q3, *localizability of important information*; Q4, *hazard differentiation*; Q5, *understanding of hazard severity*; Q6, *understanding of the surrounding environment*; and Q7, *support for evacuation decision-making*. Participants also indicated their overall map preference (Q8). For Q1–Q7, we compared the two conditions using paired Wilcoxon signed-rank tests on 7-point Likert ratings (1 = Strongly Disagree, 7 = Strongly Agree). As shown in Figure 6, maps generated by our system received significantly higher ratings for map clarity and readability (Q1, $p < .001$), information amount (Q2, $p < .001$), localizability of important information (Q3, $p < .001$), hazard differentiation (Q4, $p < .001$), understanding of the surrounding environment (Q6, $p < .001$), and support for evacuation decision-making (Q7, $p = .012$). The largest contrasts were observed for Q1–Q3, where ratings for our maps were predominantly positive while ratings for the official maps tended to be more negative. For Q4, Q6, and Q7, both conditions were generally rated positively, but participants rated our maps significantly higher. No significant difference was observed for understanding hazard severity (Q5, $p = .388$), with both map conditions receiving predominantly positive ratings. Participants also showed a strong overall preference for the maps generated by our system: 20 of 24 participants (83.3%) preferred our maps, compared with two (8.3%) who preferred the official municipality maps and two (8.3%) who expressed no preference. However, three participants who preferred the system-generated maps initially raised concerns about the reliability of AI-generated hazard maps. After we explained the system's workflow, particularly that the data is obtained from the public office and the final maps are produced through a deterministic rendering pipeline rather than direct image generation, these participants reported that the concerns were reduced. They suggested this information could also be presented to the public alongside the hazard maps.

**User Feedback.** Participants generally found the AI-generated maps clearer and easier to navigate than the official maps, particularly due to stronger emphasis on important labels and shelters, fewer confusing categories, clearer basemaps for route planning, and more salient terrain and elevation cues. At the same time, some raised concerns about whether automated information filtering might omit important details, highlighting a tension between simplification and perceived reliability. Several participants therefore viewed the two map types as complementary: simplified maps could support rapid public understanding, while official maps could remain detailed and authoritative references. Preferences also varied across participants. While most preferred the color schemes of the AI-generated maps, some favored the official maps aesthetically. Views also differed on textual explanations: some found additional instructions helpful, while others considered them unnecessary when the visuals were self-explanatory. Although the simplified basemap was generally appreciated, some participants felt that major roads were less prominent than in the official maps depicted in distinct colors, potentially limiting route planning. Opinions also differed on historical hazard information: some saw past events as less relevant to current risk, while others felt the empirical evidence made risks more tangible. Participants identified several limitations of the generated maps. Some suggested strengthening terrain shading so that topographic information could be more clearly distinguished from hazard layers. Others noted that visually subduing contextual information could make certain features, such as small creeks and local roads, harder to distinguish. Some pointed out that railways and stations should receive greater emphasis, particularly because of their relevance to everyday mobility. Looking forward, participants suggested greater adaptability, including adjustable information display for different users, map size and resolution matched to displayed content, more consistent symbolization across regions, and integration of real-time hazard conditions and accessible evacuation routes.

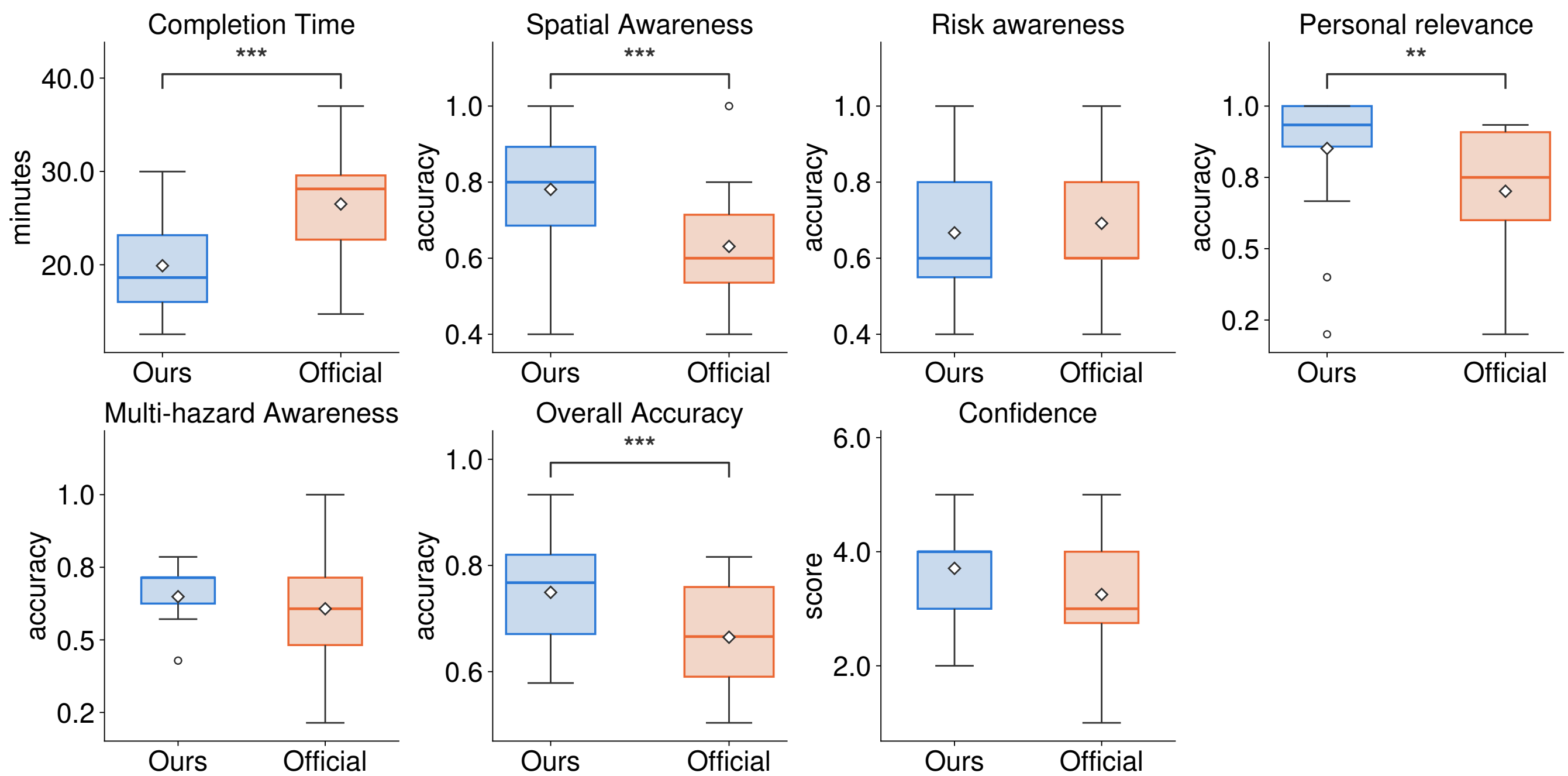

**Figure 4: Effects on hazard understanding across two map conditions: ours (AI-generated) and official (municipality). We compare the accuracy of spatial awareness, risk awareness, personal relevance, multi-hazard awareness as well as the overall accuracy and completion time. Significance levels are denoted by** * $p < .05$, ** $p < .01$, **and** *** $p < .001$.

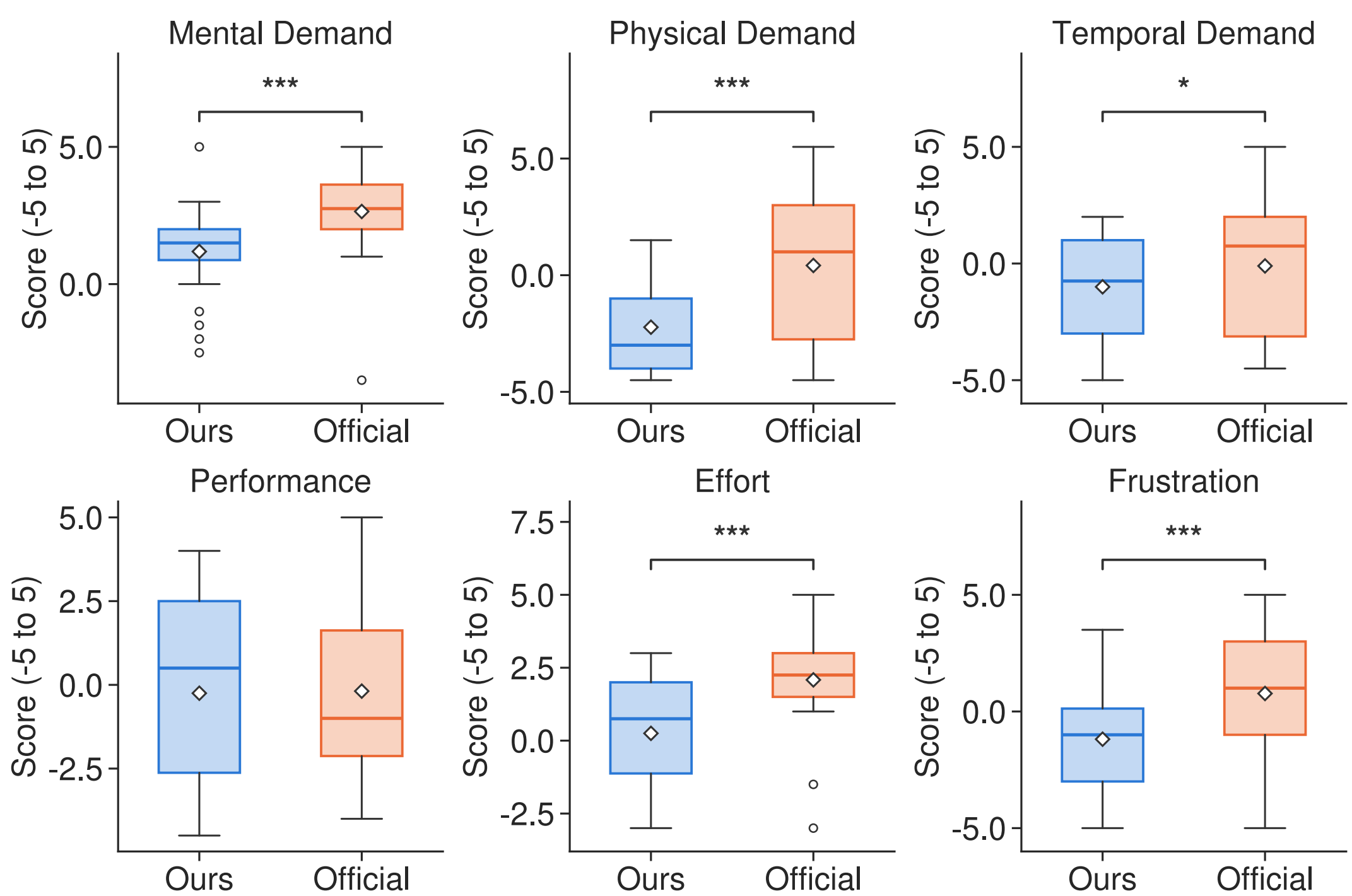

**Figure 5: Participants' perceived workload across the two map conditions, measured using NASA-TLX. Significance levels are denoted by** * $p < .05$, ** $p < .01$, **and** *** $p < .001$.

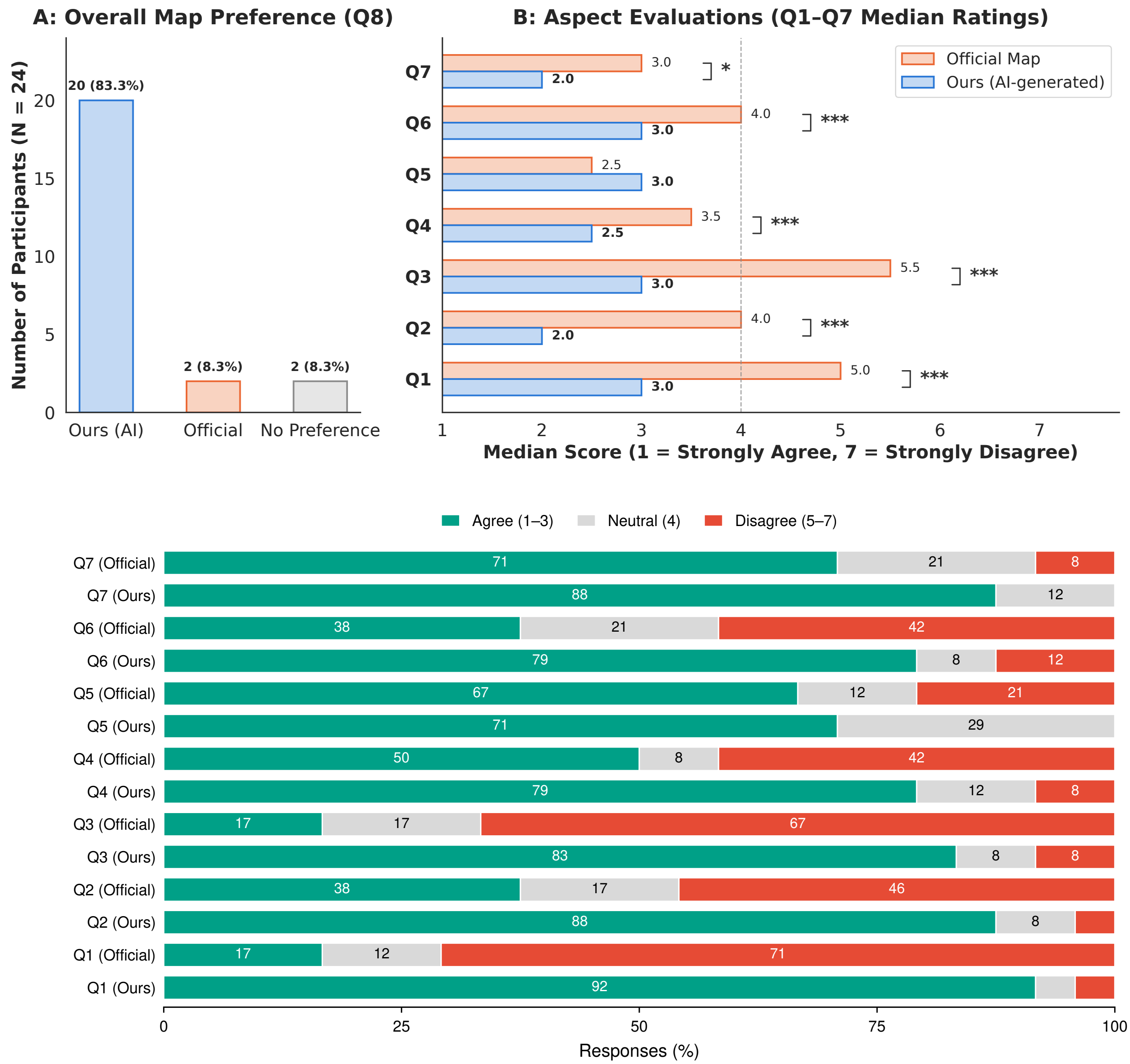

**Figure 6: User preferences across the two map conditions. (A) Overall map preference. (B) Preference ratings across seven aspects using a 7-point Likert scale (1 = Strongly Agree, 7 = Strongly Disagree). (C) Distribution of responses for each Likert-scale item. For (B) and (C), the questions assess: Q1, map clarity and readability; Q2, appropriateness of the amount of information; Q3, ease of locating important information; Q4, differentiation between hazards; Q5, understanding of hazard severity; Q6, understanding of the surrounding environment; and Q7, support for evacuation decision-making. Significance levels are denoted by * (p < .05), ** (p < .01), and *** (p < .001).**

## 6 Expert Interview

The expert interviews aimed to assess the quality of the generated maps, examine the system's practical utility within hazard-mapping workflows, and identify opportunities for further improvement. We recruited four map experts (P1–P4), who primarily focused on cartographic quality and system design, as well as one hazard practitioner (P5) and one hazard expert (P6), who primarily evaluated the system from the perspective of real-world hazard-preparedness practice.

### 6.1 Procedure.

The interviews consisted of two parts. First, we presented participants with paired examples of official municipality hazard maps and the corresponding maps generated by our system. Map experts (P1–P4) were asked to assess whether the generated maps followed established cartographic principles and to compare the two map conditions in detail, focusing on aspects such as visual hierarchy, symbolization, readability, and information organization. Hazard practitioner (P5) and expert (P6) were asked to provide a broader assessment of the generated maps, with particular attention to their usefulness and appropriateness for hazard communication and preparedness. Second, we presented the system architecture, intermediate outputs, and iterative refinement process. Map experts were asked whether the system reflected a reasonable cartographic workflow, whether the coordination and refinement mechanisms were appropriate, and how the system could be further improved. The hazard practitioner and expert were asked how, and under what conditions, the system could be integrated into practical hazard-mapping workflows, as well as what limitations or requirements would need to be addressed before real-world deployment.

### 6.2 Insights and Findings

Here we summarize the main insights from the expert interviews.

**The generated maps are consistent with cartographic principles and generally higher in quality.** All four map experts agreed that the generated maps followed key cartographic principles, including appropriate information generalization and simplification, a clear visual hierarchy that emphasized critical information, interpretable terrain representation, appropriately sized labels, and well-structured legends. When compared with the corresponding municipality maps, all four experts generally preferred the maps generated by our system. Experts also favored the system-generated color schemes for communicating relative risk levels, although their preferences varied across specific designs. For example, P2 and P3 noted that the system's tendency to use darker colors for the most severe hazard levels could make high-risk areas visually heavy or less comfortable to view. P2 additionally observed that symbols were occasionally enlarged too aggressively to establish visual hierarchy, which could obscure useful spatial information underneath. Similarly, P4 reported instances of inconsistent prioritization of overlapping hazard layers and variations in legend organization across generations, which sometimes could reduce readability and interpretability.

**The system workflow matches the cartographic practice and the iterative process is reasonable.** All four map experts agreed that the system broadly reflects their professional map-making workflow, although P2 noted that the system makes some design choices more stochastically than a human cartographer would. The evaluation feedback generated by the VLM also generally aligned with the experts' own assessments (P1–P4), supporting the rationale of the subsequent refinement process. P2 further noted that the improvements across iterations were consistent with their overall professional judgment, even though cartographers do not necessarily evaluate maps by explicitly decomposing quality into a fixed set of criteria during everyday practice. Experts particularly appreciated that the system produces an editable QGIS project through MCP rather than generating maps solely as raster images. This allows map-makers to inspect and directly modify individual map layers, symbols, labels, and other design elements when further adjustment is needed.

**Future systems could better support interpretable pictogram and symbol design.** Map experts P2 and P3 suggested that future versions of the system could provide stronger graphical support for designing symbols that are intuitive and culturally appropriate, particularly for elements such as shelters. As P3 put it, well-designed pictograms can communicate information more efficiently than text alone and may also be used in explanatory panels to reduce lengthy texts ("a picture is worth a thousand words"). Beyond symbol creation, P4 also highlighted opportunities for more sophisticated symbol manipulation and composition, such as combining symbols or deriving composite symbols through union or intersection operations. Such operations could help manage visual conflicts and reduce visual complexity, particularly in areas where multiple hazard layers overlap.

**Future systems could generate multiple map outputs adapted to different users and situations.** Given the relatively low cost and rapid generation of maps, experts suggested that future versions of the system could produce multiple outputs tailored to different needs. First, maps could be adapted to different user groups (P3). For example, land-use planners may require detailed, fine-grained information to support policy and planning decisions, whereas the general public may benefit from more simplified, generalized, and aggregated representations that support rapid evacuation decisions. Second, maps could be adapted to different spatial and situational contexts (P1, P4). For instance, a community-scale view may be more appropriate when an entire neighborhood is exposed and evacuation requires moving to another area, while a more local-scale view may be preferable when nearby safe areas remain available.

**Future systems could incorporate both expert knowledge and user preferences.** While expert preferences provide important guidance for maintaining cartographic quality, P2 suggested that future systems could also incorporate user preferences to better accommodate the needs of the general public. However, because end users may not have the same domain expertise as professional cartographers, P2 argued that user control should be more constrained than expert control. For example, if future versions support interactive specification through prompts, users could adjust high-level preferences or communication needs, while expert-derived principles and constraints continue to govern critical cartographic decisions.

**Broadening access, human oversight, and the scope of AI support.** Hazard practitioner P5 and hazard expert P6 both recognized the potential of the system to support municipalities with limited cartographic expertise. At the same time, they emphasized the importance of maintaining human involvement in AI-assisted hazard-map production. P6 particularly raised concerns about responsibility and liability for the final output, stressing the need for human supervision and accountability. P5 valued the editable QGIS output, which allows local hazard-map makers to inspect and adjust generated maps, and suggested additional LLM-based interaction for those unfamiliar with QGIS, allowing them to specify desired edits through natural language. Both P5 and P6 also envisioned broader roles for AI beyond the current system. P5 distinguished preparedness-oriented hazard communication from warnings and alerts during an emerging or ongoing emergency. While the current system focuses on preparedness maps that communicate what could happen, future systems could extend toward warning and alert communication that conveys what is happening and what actions people should take, supported by dynamic information such as real-time hazard conditions, infrastructure status, route accessibility, and recommended actions. P6 further noted that expertise across municipalities may not be limited to cartographic design, but also at earlier stages such as data collection, processing, and physical hazard modeling, suggesting opportunities for AI support across a broader hazard-mapping lifecycle. He also suggested that extending to an interactive system could support public understanding and hazard education, particularly among younger audiences.

## 7 Discussion

We now discuss about the main findings from user study and expert interviews, limitations and outlook.

### 7.1 Main Findings

**Incorporating expert cartographic knowledge can support risk communication.** By incorporating cartographic rules, quality criteria, and expert heuristics, our system can output maps that improve risk communication, particularly in spatial understanding and evacuation planning. Participants frequently attributed the usefulness of the generated maps to characteristics such as selective abstraction, information prioritization, visual hierarchy, basemap clarity, and terrain representation. These design features can help users to locate information easily, therefore support hazard understanding while reducing the effort required to interpret the maps.

**Our system can emulate expert cartographic practice.** By mimicking key stages of the cartographic workflow, the system can generate maps that satisfy core cartographic requirements while iteratively adjusting design choices to balance competing quality criteria, guided by the experts' preferences. This provides a practical foundation for map production in contexts where access to cartographic expertise is limited, while still supporting expert intervention through editable QGIS outputs enabled by MCP. More broadly, our findings suggest that agentic systems can operationalize cartographic expertise by separating hard and soft constraints, grounding in established rules while preserving the ability to adapt concrete design choices to different data, communities, and hazard contexts.

**"Less is more", but simplification depends on the use case.** While people generally preferred simplified AI-generated maps due to their ease of interpretation, they also expressed concerns that excessive abstraction could omit or suppress information relevant for decision-making. The optimal level of simplification should depend on context and the intended use case. While aggressive abstraction can support rapid comprehension for public hazard communication, fine-grained information can provide a reference for detailed assessment and planning-oriented tasks.

**"Automate the map, not the responsibility."** Although the system can automate substantial parts of municipal hazard-map production, responsibility, liability, and accountability remain important concerns for both users and practitioners. This suggests that human review and transparent provenance remain important for establishing AI-generated maps as authoritative public information.

**"No one-size-fits-all map", toward user-adaptive design.** A single representation may not satisfy all users as features considered peripheral by one group may be critical for another. Therefore, rather than universally optimized maps, user-centered personalization on map content, level of detail, scale, and design can be investigated. In the context of risk communication, aesthetic preference may matter less than interpretive needs and prior knowledge. For example, users with lower map literacy may benefit from simplified representations, explicit annotations, and familiar symbols, whereas more experienced users may prefer denser information and greater technical detail.

### 7.2 Limitations and outlook

Here we also discuss about limitations and point out to future directions.

**Covering different stages of risk communication.** Our study primarily focused on hazard preparedness and evacuation planning rather than near-disaster warnings or emergent actions during an ongoing disaster. This setting allowed us to examine how users interpret and plan with AI-generated maps under relatively stable information conditions, but does not capture the temporal pressure and rapidly changing information characteristic of an unfolding emergency. But since our system can output maps rapidly at relative low expense, future work could extend it across different stages of disaster management, from preparedness and early warning to active evacuation, by incorporating real-time hazard conditions, infrastructure status, traffic conditions, and accessible evacuation routes, paired with textual warning, alerts, and other modalities.

**Learning implicit cartographic preference.** The current map-quality scoring function was derived from a limited number of experts and relied primarily on explicit and fixed preference weights. In practice, however, cartographic preferences are often contextual, implicit, and difficult to fully articulate. Experts may make different design decisions depending on the task, audience, geographic setting, and communication goals. Future work could investigate preference-alignment approaches that learn expert preferences from ranked pairs of map designs [31] and larger collections of cartographic examples, enabling more flexible and context-sensitive cartographic reasoning.

**Enabling a broader cartographic design space.** The current system provides relatively limited functionality of manipulating map features and design parameters, primarily focusing on information selection and prioritization, colors schemes, and symbol and font sizes. Future systems could broaden the design space by enabling the generation of pictogram or symbols that are intuitive within specific cultural context or supporting more complex symbol operation, to better support visual communication and storytelling. Additional map-generalization functions, such as shape simplification and aggregation could further improve the communication of complex spatial information [2, 14]. Beyond symbolization, future systems could also support more advanced layout design by intelligently arranging map, legend, and explanatory panels, allowing greater flexibility in the organization and presentation of information.

**Extending to other hazard types and locations.** The current study focused on flood and sediment hazards. Despite variation in their mechanisms, impacts, and interactions, the findings may not generalize to other hazard types of different spatial and temporal attributes. Future work could investigate hazards such as tsunamis, earthquakes, wildfires, and extreme weather events, where near-real-time information, uncertainty communication, and rapidly evolving conditions play a more central role. Furthermore, due to data availability constraints, the system was evaluated in a limited number of cities. Additional studies are needed to assess its generalizability across diverse geographic settings, data conditions, and community needs.

**Supporting local map makers through accessible interaction.** For maps intended for official use, municipal map makers may still need to inspect and revise AI-generated outputs before publication. Although the current system produces editable QGIS projects that support further manual refinement, directly editing these outputs still requires familiarity with the software. This may limit accessibility for municipal staff responsible for hazard-map production who possess local and hazard-domain knowledge but limited Geographic Information System (GIS) or cartographic expertise. Future systems could therefore support natural-language interaction, allowing local map makers to inspect the generated design, specify desired changes, and modify the map without directly manipulating complex GIS interfaces. Such interaction could retain human oversight while lowering the technical barrier to adapting generated maps to local needs.

**Extending AI support across the hazard-mapping lifecycle.** The current system focuses primarily on cartographic production, although expertise may also be limited at earlier stages such as data collection, processing, and physical hazard modeling. Future work could investigate how AI support can extend across this broader lifecycle through a more comprehensive agentic framework that coordinates specialized agents across different stages.

**Adapting to users, situations, and communication needs.** Beyond supporting map makers in reviewing and approving a map for official use, future systems could also adapt the resulting representations to different audiences and communication contexts. Importantly, such adaptation should concern how authoritative hazard information is communicated rather than altering the underlying hazard data, categories, or meanings. Beyond static printed maps, interactive systems could adapt map content and presentation according to users' interpretive needs, prior knowledge, local community contexts, and personal preferences, while remaining grounded in expert cartographic guidance and official hazard-mapping requirements.

## 8 Conclusion

We presented *Agentic Hazard Mapper*, a multi-agent system for automated hazard-map production that emulates key stages of professional cartographic practice and iteratively refines its outputs through evaluation and revision. Grounded in cartographic rules, hazard-mapping requirements, and expert-derived practices, the system coordinates specialized agents for data processing, visual design, layout, evaluation, and refinement while producing editable outputs through QGIS. Across a formative study, user study, and expert interviews, our findings show that incorporating cartographic expertise into an agentic workflow can improve the effectiveness of hazard maps for risk communication. Compared with corresponding municipality maps, the generated maps supported higher task accuracy and faster completion, reduced perceived workload, and were generally preferred by both users and map experts. Participants particularly valued clearer information prioritization, stronger visual hierarchy, improved basemap readability, and more interpretable terrain representation. At the same time, our studies highlighted trade-offs between simplification and informational completeness, differences in needs across users and communication contexts, and the continued importance of human oversight for maps intended for official use. These findings suggest that the goal of automated hazard-map production should not be to produce a single universally optimal map, but to make cartographic expertise more accessible while enabling context-sensitive and user-adaptive representations within authoritative hazard-mapping constraints. More broadly, our work points toward agentic systems as a promising approach for supporting complex visual-communication tasks that require both domain expertise and flexible, iterative design judgment.